\documentclass[preprint]{elsarticle}
\usepackage{amssymb}
\usepackage{amsmath}
\usepackage{subcaption}
\usepackage{array}
\usepackage{multirow}
\usepackage[table]{xcolor}
\usepackage{url}
\makeatletter
\def\ps@pprintTitle{%
   \let\@oddhead\@empty
   \let\@evenhead\@empty
   \let\@oddfoot\@empty
   \let\@evenfoot\@oddfoot
}
\makeatother

\begin{document}

\begin{frontmatter}

\title{Zero-Shot LLMs in Human-in-the-Loop RL: Replacing Human Feedback for Reward Shaping\tnoteref{label1}} 

\author[1]{Mohammad Saif Nazir}
\ead{mohammad_2303res105@iitp.ac.in}

\author[2]{Chayan Banerjee\corref{cor1}}
\ead{c.banerjee@qut.edu.au}

\cortext[cor1]{Corresponding author.}

\affiliation[1]{
  organization={IIT Patna}, 
  addressline={Bihta Kanpa Rd.},
  city={Patna}, 
  postcode={801106}, 
  state={Bihar}, 
  country={India}
}

\affiliation[2]{
  organization={Queensland University of Technology}, 
  addressline={2 George Street},
  city={Brisbane}, 
  postcode={4000},
  state={Queensland}, 
  country={Australia}
}

\begin{abstract}
Reinforcement learning (RL) often struggles with reward misalignment, where agents optimize given rewards but fail to exhibit the desired behaviors. This arises when the reward function incentivizes proxy behaviors misaligned with the true objective. While human-in-the-loop (HITL) methods can mitigate this issue, they also introduce biases, leading to inconsistent and subjective feedback that complicates learning.
To address these challenges, we propose two key contributions. First, we extend the use of zero-shot, off-the-shelf large language models (LLMs) for reward shaping beyond natural language processing (NLP) to continuous control tasks. Using LLMs as direct feedback providers eliminates the need for surrogate models trained on human feedback, which often inherit biases from training data. Second, we introduce a hybrid framework (LLM-HFBF) that enables LLMs to identify and correct biases in human feedback while incorporating this feedback into the reward shaping process.
The LLM-HFBF framework creates a more balanced and reliable system by addressing both the limitations of LLMs (e.g., lack of domain-specific knowledge) and human supervision (e.g., inherent biases). By enabling human feedback bias flagging and correction, our approach improves reinforcement learning performance and reduces reliance on potentially biased human feedback.\newline
Empirical experiments show that biased human feedback significantly reduces performance, with Average Episodic Reward dropping by nearly 94\% compared to unbiased approaches. In contrast, LLM-based methods sustain performance at a similar level to unbiased feedback, even in challenging edge-case scenarios.

\end{abstract}

\begin{keyword}
Zero-Shot LLMs \sep Human-in-the-Loop \sep Reinforcement Learning \sep Reward Shaping \sep Bias correction
\end{keyword}

\end{frontmatter}

\section{Introduction}
Reinforcement learning is a quantified process of exploring and exploiting an environment to maximize a numerical reward signal by discovering efficient actions \cite{sutton1998reinforcement}. However, the design of effective reward signals often demands extensive domain expertise and iterative tuning. The process of defining the reward function, often suffers from \textit{reward misalignment} where the agent may devise a way to boost rewards without genuinely achieving the desired task outcome \cite{krakovna2020avoiding}.\newline
For example in the game \textit{CoastRunners}, while players are expected to complete a boat race quickly, the reward system prioritizes hitting targets along the course rather than finishing the race itself. When trained via RL, the agent learned to maximize its score by circling an isolated lagoon and repeatedly hitting three respawning targets, thus exploiting the reward structure without completing the race. This demonstrates how reward misalignment can lead to unintended behaviors, where the agent's strategy diverges from the designer's true intentions \cite{openai_faulty_reward}.

Human-in-the-loop reinforcement learning (HITL-RL) has garnered attention as a potential solution to this problem \cite{saunders2017trial}. It incorporates human feedback into the RL process to refine or even directly shape the reward function. For instance, \cite{christiano2017deep} demonstrated that leveraging human feedback can guide RL agents more effectively toward desirable behaviors, mitigating issues like reward hacking and misaligned incentives. 

But despite the success of human-in-the-loop methods, their heavy reliance on human intervention presents several challenges, particularly limiting scalability and efficiency. The high cost of human feedback, difficulties in providing consistent feedback in real-time, and scalability issues when handling large, continuous state-action spaces further complicate these methods. Most importantly human biases can interfere with the agent's ability to generalize across diverse situations \cite{casper2023open}. 

\begin{figure}
    \centering
    \includegraphics[width=0.8\linewidth]{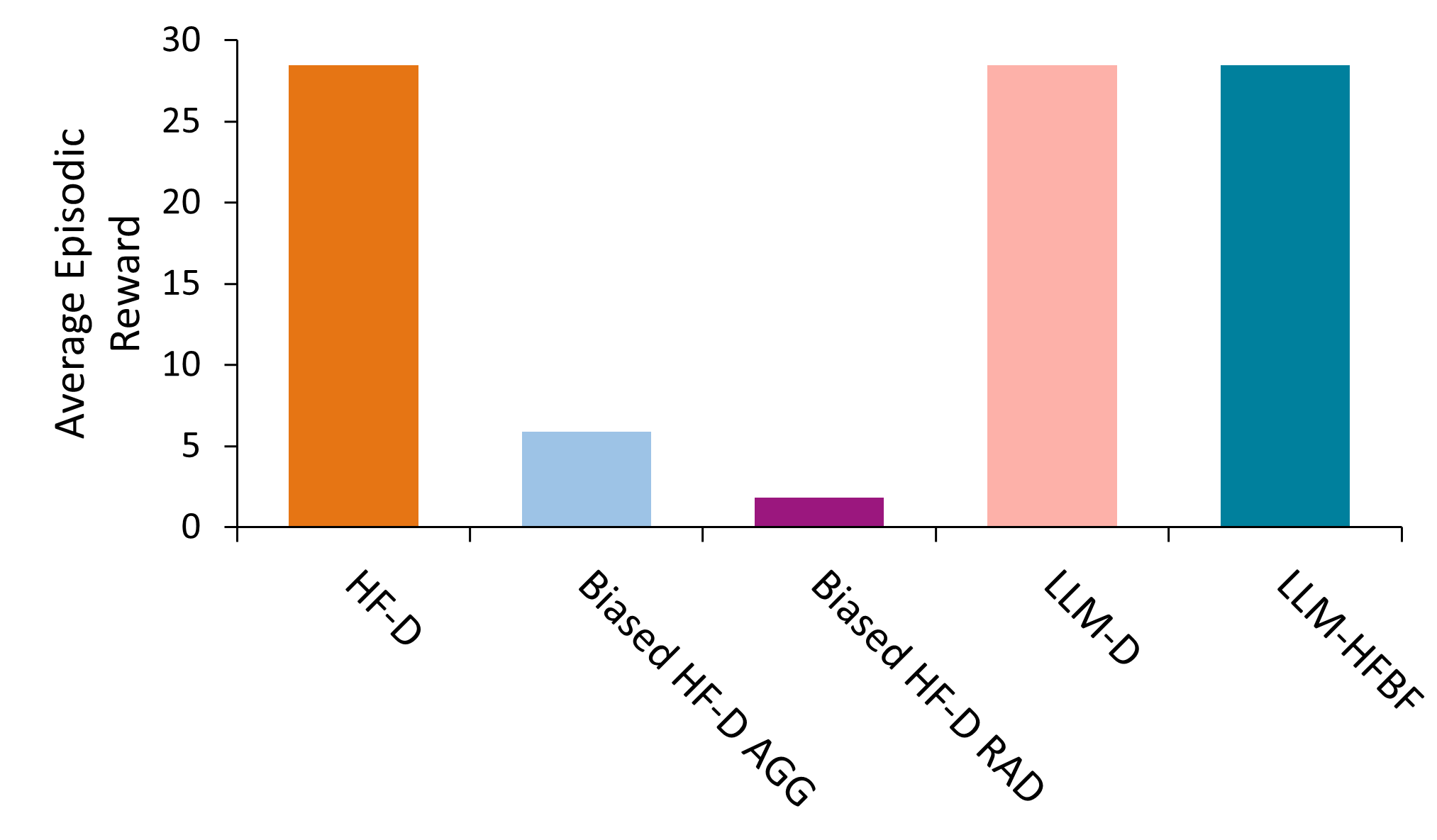}
    \caption{Performance comparison of direct human feedback (HF-D), biased human feedback (Biased HF-D(AGG) and Biased HF-D(RAD)), and \textit{proposed LLM-based feedback methods (LLM-D and LLM-HFBF)} for reward shaping and RL training. Results are from Highway-v0 environment. }
    \label{fig:first_img}
\end{figure}
Personal human biases and preferences can distort feedback, preventing optimal learning \cite{christiano2017deep}. 
Several approaches have tried to train and use surrogate models as a proxy to direct human feedback to make HITL-RL more cost effective, for example Reinforcement Learning from AI Feedback (RLAIF)  \cite{leerlaif} and Iterative Reward Shaping (ITERS) \cite{gajcin2023iterative}. However, these methods face significant limitations: ITERS follows human feedback without accounting for potential biases, while RLAIF relies entirely on LLM-generated feedback which may lack the nuanced human understanding required in certain complex scenarios. Both approaches can be compromised when the feedback itself suffers from human bias, as the surrogate models inevitably inherit and potentially amplify these biases during training.\newline

While LLMs can inherit biases from their pretraining corpora~\cite{doi:10.1073/pnas.2313790120}, including behavioral and social biases such as ingroup favoritism~\cite{hu2024generativelanguagemodelsexhibit} and systematic biases in opinion formation exploitable through adversarial attacks~\cite{rennard2024biasmirrorllmsopinions}, they also offer distinct advantages over human feedback in reinforcement learning contexts. Human feedback is often inconsistent, subjective, and variable over time, whereas LLM biases tend to be systematic and therefore potentially auditable. Although alignment techniques such as instruction tuning and preference modeling do not fully eliminate LLM biases~\cite{hu2024generativelanguagemodelsexhibit}, they establish a foundation for generating feedback that is more predictable and scalable.


In this work, we first demonstrate that off-the-shelf LLMs can effectively shape rewards in continuous control tasks within the MuJoCo environment, requiring no fine-tuning (zero-shot). However, recognizing that human feedback remains essential in certain scenarios—particularly in high-risk and sensitive environments—we introduce a hybrid framework where LLMs identify potential biases in human feedback, enabling targeted bias correction. To illustrate the impact of different feedback strategies, Figure \ref{fig:first_img} compares the performance of direct human feedback (HF-D), human feedback influenced by aggressive and reckless adaptive biases (Biased HF-D(AGG) and Biased HF-D(RAD), respectively), and \textit{our proposed LLM-based methods}: a direct reward shaping approach (LLM-D) and a hybrid feedback framework that flags and corrects biases (LLM-HFBF). 
The code associated with this paper is publicly available\footnote{\url{https://github.com/RizanSM/zero_shot_llms_in_HIL_RL}}.\newline

The main contributions of this paper are as follows:
\begin{itemize}
\item Through empirical experiments in simulated environments, we show that even minor biases in human feedback can significantly degrade reward shaping quality, leading to substantial performance drops in learned policies.
\item We introduce the use of off-the-shelf LLMs as zero-shot feedback providers (in the context of continuous control) to replace potentially biased human feedback in HITL-RL, avoiding the costs of continuous human feedback and the risk of inherited biases from learned reward models.
\item We validate our approach in continuous control tasks (\textit{Highway-v0} and \textit{Reacher-v5 } in MuJoCo), demonstrating that zero-shot LLMs can effectively shape rewards in these domains at par their unbiased counterparts.
\item We propose a hybrid framework  where LLMs not only provide feedback but also flag human biases, facilitating bias correction in scenarios where human input remains valuable.
\end{itemize}

\section{Related work}
\noindent
Human-in-the-loop RL (HITL-RL) has been explored as a solution to reward misspecification, where human feedback refines or directly shapes reward functions \cite{saunders2017trial}. While methods such as learning from human preferences \cite{christiano2017deep} simplify reward design, they suffer from high annotation costs, inefficient exploration, and uncertainty in learned reward models. Augmenting preference learning with expert demonstrations \cite{ibarz2018reward} helps improve exploration but introduces dependencies on demonstration quality and increases computational overhead.

Other approaches have focused on real-time human feedback, such as COACH \cite{macglashan2017interactive} and Deep TAMER \cite{warnell2018deep}, which iteratively refine policies using human signals. However, these methods face scalability challenges due to their reliance on continuous human involvement. Adaptive shaping \cite{yu2018learning} and offline human feedback approaches like FRESH \cite{xiao2020fresh} address some of these limitations, but they struggle with dynamic environments, computational overhead, and adaptability.

Recent methods such as EXPAND \cite{guan2021widening} and RL-TF \cite{efroni2021reinforcement} attempt to improve efficiency by integrating visual explanations and trajectory-level feedback, respectively. However, EXPAND increases human workload, and RL-TF introduces estimation uncertainty. These approaches highlight the ongoing challenge of balancing scalability, accuracy, and bias mitigation in human feedback-driven RL.

Cognitive and contextual biases in human feedback pose significant risks to RL policy optimization \cite{casper2023open}. Feedback inconsistency, subjectivity, and over-reliance on prior interactions can reinforce biased behaviors in RL agents. Studies show that feedback loops between humans and AI can amplify biases, as humans often underestimate AI influence \cite{glickman2024human}. Additionally, simplifications in existing HITL-RL models fail to capture the nuances of human decision-making, leading to suboptimal performance in real-world applications \cite{lindner2022humans}.

To reduce reliance on direct human annotations, many HITL-RL methods incorporate surrogate models that learn from human feedback to approximate reward functions \cite{gajcin2023iterative}. The ITERS framework \cite{gajcin2023iterative} follows this approach by iteratively refining a reward shaping model using human-labeled trajectories. However, these models risk bias amplification, as errors in initial feedback propagate through the learning process. Additionally, training a surrogate requires an augmented dataset of trajectories, introducing computational overhead and reliance on feedback quality.

Beyond ITERS, other surrogate-based approaches such as fitness estimation via phenotypic distance \cite{Stork_2019} and scalable high-dimensional modeling \cite{stapleton2024neurolgp} have been explored. However, these methods still inherit biases from human-labeled data and struggle with generalization across diverse RL tasks.

Efforts to reduce reliance on human feedback include synthetic data generation and AI-driven annotations. While early AI-generated feedback techniques show promise, they lack robustness in complex tasks and do not fully address bias correction. These limitations have motivated research into leveraging Large Language Models (LLMs) as feedback providers.

LLMs have recently been used in RL for reward design and preference modeling \cite{cao2024enhancing, chan2024dense}. Methods such as step-wise critique and dynamic reward aggregation improve training efficiency by generating intermediate-step rewards, reducing reward sparsity. However, these approaches often depend on the quality of the pre-trained models and may not generalize across tasks.

RLAIF (Reinforcement Learning from AI Feedback) \cite{leerlaif}  directly replaces human feedback with LLMs as zero-shot evaluators, enabling scalable AI-driven reward shaping. While effective for NLP applications (e.g., text summarization, dialogue generation, and question answering), RLAIF has not been extensively applied to continuous control tasks. Moreover, existing LLM-based methods do not explicitly address bias detection and correction, a crucial factor in HITL-RL systems.

Bias mitigation techniques in RL have focused on uncertainty quantification and reward model adjustments. However, most existing work prioritizes improving reward models rather than identifying and correcting biased feedback in real-time \cite{gao2023off}. Ensuring unbiased feedback remains a key challenge for AI-driven reward shaping.

\vspace{0.2cm}
\noindent
Our work differs from prior approaches in three key ways:\newline
\noindent
- Zero-Shot LLMs for Continuous Control: Unlike RLAIF\cite{leerlaif}, which primarily targets NLP applications, we extend zero-shot LLM-based reward shaping to continuous control tasks in MuJoco Environment.\newline
- Avoiding Surrogate Models Susceptible to Bias: Unlike ITERS \cite{gajcin2023iterative}, which learns a reward shaping model from human feedback, we use off-the-shelf LLMs to provide structured, unbiased feedback directly, avoiding bias amplification risks.\newline
- Hybrid Framework for Bias Detection and Correction: Recognizing that certain problems may require unbiased human feedback due to their unique nature, we propose a two-stage approach in which LLMs not only provide reward shaping but also identify and flag potential biases in human feedback.

\section{Preliminaries}
\noindent Reinforcement Learning (RL) is a sequential decision-making framework where an agent interacts with an environment to maximize cumulative rewards. It is formalized as a Markov Decision Process (MDP), defined by the tuple \( (S, A, P, R, \gamma) \), where \( S \) is the state space, \( A \) is the action space, \( P: S \times A \times S \to [0,1] \) represents transition probabilities, \( R: S \times A \times S \to \mathbb{R} \) is the reward function, and \( \gamma \in [0,1] \) is the discount factor. The agent follows a parameterized policy \( \pi_{\phi}(a \mid s) \), aiming to maximize the expected cumulative reward:

\begin{equation}
J(\pi_{\phi}) = \mathbb{E}_{\tau \sim \pi_{\phi}} \left[ \sum_{t=0}^{\infty} \gamma^t r_t \right]
\end{equation}

At each time step \( t \), the reward signal \( r_t \) is obtained from the reward function:

\begin{equation}
r_t = R(s_t, a_t, s_{t+1})
\end{equation}

To enhance learning, the reward shaping approach introduces a shaped reward $\hat{r}_t$, resulting in an augmented reward given by

\begin{equation}
\tilde{r}_t = r_t + \hat{r}_t
\end{equation}

modifying the objective to

\begin{equation}
J(\pi_{\phi}) = \mathbb{E}_{\tau \sim \pi_{\phi}} \left[ \sum_{t=0}^{\infty} \gamma^t \tilde{r}_t \right].
\end{equation}

Policy optimization, often using methods like Proximal Policy Optimization (PPO)\cite{schulman2017proximal}, leverages the policy gradient theorem:

\begin{equation}
\nabla_{\phi} J(\pi_{\phi}) = \mathbb{E}_{\tau \sim \pi_{\phi}} \left[ \sum_{t=0}^{\infty} \nabla_{\phi} \log \pi_{\phi}(a_t \mid s_t) G_t \right]
\end{equation}

where \( G_t \) is the advantage function estimating expected future rewards.

\begin{figure}[htbp]
    \centering
    \includegraphics[width=0.95\linewidth]{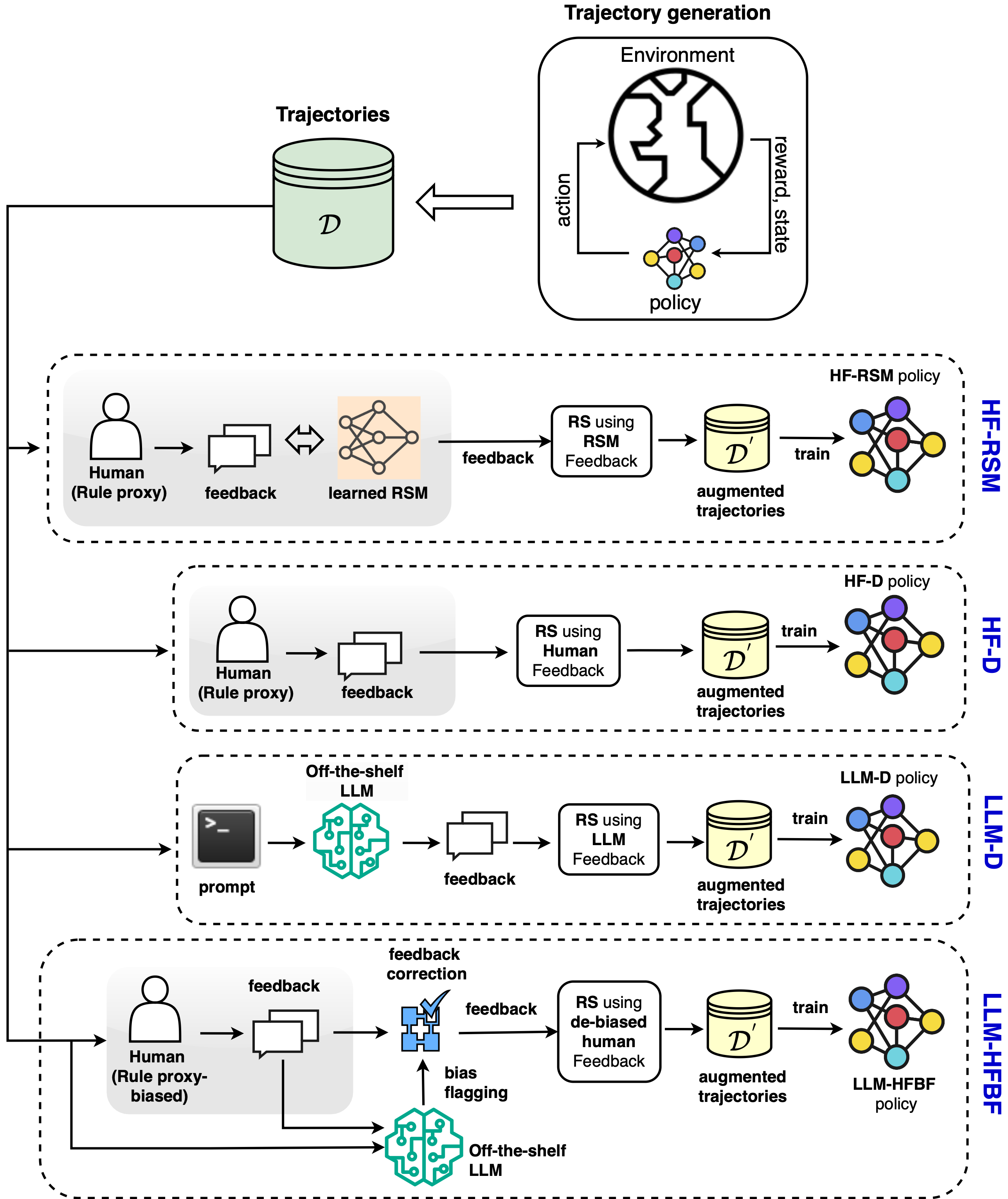}
    \caption{Feedback integration strategies (FIS). HF-RSM: Human Feedback using learned Reward Shaping Model, HF-D: Human Feedback used directly, LLM-D: LLM suggested feedback used directly, LLM-HFBF: LLM used for Human Feedback Bias flagging, which is utilized for human feedback debiasing. }
    \label{fig:feedback_strategies}
\end{figure}

\section{Methodology}
\label{sec:methodology}
To investigate bias-induced performance degradation in feedback systems, particularly within the context of reward shaping, and to assess their application with large language models (LLMs), we implement three distinct behavioral profiles using rule-based proxies: IDEAL (balanced), AGGRESSIVE (risk-seeking), and RECKLESS ADAPTIVE (impulsive opportunistic). Each profile is characterized by style coefficients that influence lane changes, collision avoidance, and speed optimization. We compare four feedback integration strategies: HF-RSM (human feedback with learned reward shaping model), HF-D (direct human feedback), LLM-D (direct LLM feedback), and LLM-HFBF (hybrid approach where LLMs detect and adapt to human feedback biases). Additionally, we design custom edge case scenarios to evaluate how these strategies handle situations where rule-based systems struggle with contextual adaptation. 

\subsection{Feedback Integration Strategies (FIS)}
\subsubsection{Overview}
In this work, we explore four primary feedback integration strategies (see Fig.~\ref{fig:feedback_strategies}) for reward shaping. They are: \textit{Human Feedback - Learned Reward Shaping Model} (HF-RSM) which represents the ITERS \cite{gajcin2023iterative} work, \textit{Human Feedback - Direct} (HF-D) same as ITERS but without the RSM training, \textit{Large Language Model - Direct} (LLM-D) works on the lines of RLAIF, and \textit{Human Feedback Bias Flagging} (LLM-HFBF) which is our proposed hybrid merging direct human and LLM feedback, with bias flagging and mitigation. \newline
All four approaches start with a set of trajectories $\mathcal{D}$ generated by running an arbitrary RL policy \(\pi_{\phi}\), for \textit{T} timesteps. The reward shaping process is then applied using one of the following feedback integration strategies: human-based feedback (HF-D, HF-RSM), direct LLM-based feedback (LLM-D), or the hybrid LLM-HFBF approach, which results in augmenting the reward value of each of these trajectories. The augmented trajectories $\mathcal{D}'$ are subsequently used to train a separate, final learned policy, \( \pi_{\psi} \), which is then evaluated.

Let \( \mathcal{D} = \{ (s_t, a_t, r_t, \hat{r}_t) \}_{t=0}^{T} \) be the trajectory set from an RL policy \( \pi_{\phi} \), where \( s_t \), \( a_t \), \( r_t \), and \( \hat{r}_t \) represent the state, action, reward, and shaped reward at time \( t \), respectively, and \( T \) is the total number of timesteps.  
Let the augmented reward be \( \tilde{r}_t = r_t + \hat{r}_t \), and the augmented trajectory:

\[
\mathcal{D}^{'} = \{ (s_t, a_t, \tilde{r}_t) \}_{t=0}^{T}
\]

The RL policy is updated using an offline learning objective based on the augmented dataset \( \mathcal{D}' \). The objective function for offline learning is typically given by:

\[
\mathcal{L}(\pi_{\psi}) = \mathbb{E}_{\mathcal{D}^{'}} \left[ \min \left( r_t(\pi_{\psi}) \hat{A}_t, \text{clip}(r_t(\pi_{\psi}), 1 - \epsilon, 1 + \epsilon) \hat{A}_t \right) \right]
\]

where \( r_t(\pi_{\psi}) = \frac{\pi_{\psi}(a_t \mid s_t)}{\pi_{\phi}(a_t \mid s_t)} \) is the probability ratio between the learned policy \( \pi_{\psi} \) and the old policy \( \pi_{\phi} \), \( \hat{A}_t \) is the \textit{advantage function}, and \( \epsilon \) is a small hyperparameter that ensures the policy update is not too large. Here, the expectation is taken over the offline dataset \( \mathcal{D}' \), and the PPO objective is adjusted to accommodate offline data.

The policy update step is given by:

\[
\pi_{\psi} = \text{Update}(\pi_{\phi}, \mathcal{D}^{'}, \mathcal{L}(\pi_{\psi}))
\]

Finally, the resulting learned policy \( \pi_{\psi} \) is deployed and tested in the environment, where its performance is evaluated based on the feedback-shaping strategies applied during training. This policy \( \pi_{\psi} \) incorporates the augmented reward signals and represents the final model for evaluation and deployment.

\subsubsection{Human Feedback - Learned Reward Shaping Model (HF-RSM)}

This approach builds upon the ITERS strategy \cite{gajcin2023iterative}, where human feedback is progressively accumulated and leveraged to train a surrogate model for reward shaping based on human feedback.
Let \( \zeta_{\text{HF-RSM}}(s; \eta) \) represent the human feedback-based reward shaping function for state \( s \), where \( \eta \) denotes the parameters of the learned surrogate model. The augmented reward at time step \( t \) is:

\[
\tilde{r}_t = r_t +  \zeta_{\text{HF-RSM}}(s_t; \eta)
\]

\subsubsection{Human Feedback - Direct (HF-D)}\label{Sec: HF-D}
Here the rule-based proxy is used directly without the surrogate model.
Although rule-based systems may not be ideal for most cases, they are well suited for our purpose, where we address the challenges of using direct human feedback in reward shaping scenarios, particularly the risk of bias in such feedback. \\
In this proof-of-concept, we use rules as a stand-in for human feedback, a common approach in related research. Rule-based systems offer transparent, consistent, and controlled decision-making, making them ideal when human feedback is limited or costly \cite{leusmanncomparing}. In RL, predefined rules approximate human preferences in the absence of direct feedback, enabling structured policy learning \cite{urcelay2024reinforcement}. Similarly, rule-based dialogue systems ensure predictable interactions in human-AI collaboration, fostering usability and trust \cite{verma2023preference}.\newline

In the Highway environment (see Sec.~\ref{Sec:Env_def}), drawing motivation from the rules in \cite{gajcin2023iterative}, we implemented three behavioral profiles to simulate human feedback: \textbf{IDEAL} (default), \textbf{AGGRESSIVE (AGG)}, and \textbf{RECKLESS ADAPTIVE (RAD)}. The \textbf{AGG} and \textbf{RAD} profiles serve as \textbf{biased feedback proxies}, with AGG encouraging risk-seeking behaviors such as rapid lane changes and higher speeds, and RAD promoting impulsive and opportunistic driving through abrupt maneuvers and inconsistent speed choices. These profiles are empirically grounded in patterns of risky and erratic driving identified in \cite{TAUBMANBENARI2016179}, which link psychological traits to driving behaviors. Each biased profile is computationally simulated by modifying the \textit{style coefficients} $B$, thereby affecting the resulting $\zeta_{\text{HF-D}}$.

\begin{align*}
\zeta_{\text{HF-D}}(p) &= w_{\text{lane}} R_{\text{lane}}(n, p) 
+ w_{\text{collision}} R_{\text{collision}}(ttc, acc, lc, p) \\
&\quad + w_{\text{speed}} R_{\text{speed}}(\rho, v, v_{\text{thresh}}, p) \\
R_{\text{lane}}(n, p) &= B_{\text{lane}}^p[n] \\
R_{\text{collision}}(ttc, acc, lc, p) &= B_{\text{collision}}^p[\text{scenario}(ttc, acc, lc)] \\
R_{\text{speed}}(\rho, v, v_{\text{thresh}}, p) &= B_{\text{speed}}^p[3(\rho-1) + \text{speed\_level}(v, v_{\text{thresh}})]
\end{align*}

\noindent where \( p \in \{ \text{IDEAL}, \text{AGG}, \text{RAD} \} \) is the parameter set, and  
\( B_{\text{lane}}^p \), \( B_{\text{collision}}^p \), and \( B_{\text{speed}}^p \) are the \textit{style coefficient} arrays for parameter set \( p \). Moreover $w_{lane}$, $w_{collision}$, and $w_{speed}$ are weights (default=1).\newline

\noindent
\textbf{Note on Scenario Indices and Parameters:} The table indices map to scenarios defined by several parameters: time-to-collision ($ttc$), acceleration ($acc$), lane change flag ($lc$), traffic density ($\rho$), agent speed ($v$), and speed thresholds ($v_{thresh}$). For Lane Changes, $n_0$-$n_3$ represent 0-3 lane changes. For Collision Avoidance, $c_0$-$c_4$ map to: potential collision ($0.5 \leq ttc \leq 2.0$) with $acc < 0$ ($c_0$), $acc > 0$ ($c_1$), or $lc = 1$ ($c_2$); immediate risk ($ttc < 0.5$, $c_3$); and safe path ($ttc > 2.0$, $c_4$). For Speed Optimization, $s_0$-$s_8$ combine traffic density levels ($\rho \in \{1,2,3\}$) with speed categories relative to thresholds: $v > v_{thresh}[1]$ (high), $v_{thresh}[0] \leq v \leq v_{thresh}[1]$ (medium), and $v < v_{thresh}[0]$ (low).

\begin{table}[htbp]
\centering
\caption{Style Coefficient  Values for Feedback representing different Driving Styles}
\scalebox{0.6}{
\begin{tabular}{|l|cccc|ccccc|ccccccccc|}
\hline
\multirow{2}{*}{\textbf{Style}} & \multicolumn{4}{c|}{\textbf{Lane Changes} $B_{\text{lane}}^p$} & \multicolumn{5}{c|}{\textbf{Collision Avoidance} $B_{\text{collision}}^p$} & \multicolumn{9}{c|}{\textbf{Speed Optimization} $B_{\text{speed}}^p$} \\
& $n_0$ & $n_1$ & $n_2$ & $n_3$ & $c_0$ & $c_1$ & $c_2$ & $c_3$ & $c_4$ & $s_0$ & $s_1$ & $s_2$ & $s_3$ & $s_4$ & $s_5$ & $s_6$ & $s_7$ & $s_8$ \\
\hline
$IDEAL$ & +1& 0 & -1 & -1& +1 & 0 & +1 & -2 & +2 & +2 & +1 & -1 & -1 & +2 & -1 & -2 & -1 & +2 \\
$AGG$ & -2 & +1 & +2 & +2 & -2 & +2 & +1 & -1 & -2 & +2 & +1 & -2 & +1 & 0 & -1 & +2 & +1 & -2 \\
$RAD$& -2 & 0& +1& +1& -2 & +1& +1 & 0& -2& +1& 0& -2 & 0& -1& -2& +1& 0& -2\\
\hline
\end{tabular}}
\vspace{0.1cm}
\caption*{This table presents the \textit{style coefficient} values across three different driving profiles: \textbf{IDEAL}, \textbf{AGG}, and \textbf{RAD}. The matrix is organized into three behavioral categories: \textit{Lane Change behavior} (\(B_{\text{lane}}^p\), based on number of lane changes \(n\)), \textit{Collision Avoidance strategies} (\(B_{\text{collision}}^p\), based on time-to-collision \(t_{\text{tc}}\), acceleration \(a_{\text{cc}}\), and lane change flag \(l_{\text{c}}\)), and \textit{Speed Optimization} (\(B_{\text{speed}}^p\), based on traffic density \(\rho\) and agent speed \(v\) relative to threshold speeds \(v_{\text{thresh}}\)). Each entry represents the style coefficient value for a specific scenario, with positive values encouraging behaviors and negative values discouraging them.
}
\label{tab:style_coefficient_matrix}
\end{table}
\subsubsection*{Human feedback variability: }
\noindent
We defined distinct behavioral subgroups to simulate varying risk tolerance and preferences. They are designed to reflect different human preferences. 
\begin{itemize}
    \item \textbf{Aggressive Feedback:} Encourages risky behaviors such as frequent lane changes and speeding up in potential collisions. It penalizes staying in the same lane and slowing down in risky situations, promoting fast-paced, assertive driving, even in high-traffic conditions. This behavior is implemented by modifying style coefficients, such as penalizing lane changes with negative values (e.g., \( B_{\text{lane}}[n_0, n_1, n_2, n_3] \)) and rewarding high speeds with positive coefficients (e.g., \( B_{\text{speed}}[s_3, s_4] \) for high-speed categories).

    \item \textbf{Reckless Adaptive Feedback (RAD):} Promotes impulsive and opportunistic driving by dynamically rewarding abrupt maneuvers and inconsistent speed choices in varying traffic conditions. The strategy prioritizes short-term gains with minimal regard for potential hazards. Safe behaviors, such as cautious path selection or slowing down to avoid collisions, are penalized, while risky actions—including sudden accelerations and frequent multi-lane changes—are rewarded regardless of safety. This behavior is implemented by adjusting the style coefficients \(B\): coefficients corresponding to safe actions (e.g., \(B_{\text{collision}}[c_0, c_4]\) for cautious path selection and slowing down) are assigned negative values, whereas coefficients associated with abrupt, high-risk maneuvers are positively reinforced.
    
\end{itemize}
%
\begin{table}[h!]
    \normalsize
    \begin{minipage}{0.48\textwidth}
        \scalebox{0.55}{
        \renewcommand{\arraystretch}{1.3}
        
        \begin{tabular}{|p{11cm}|}
            \hline
            \rowcolor{cyan!30}
            \textbf{Prompt $\rightarrow$ LLM} \\ 
            \hline
            \textbf{Data Details: } \\
            \textbf{Episode:} \texttt{\{episode\_num\}} \\
            \textbf{Time Step:} \texttt{\{time\_step\}} \\
            \textbf{State:} \texttt{\{state\}} \\
            \textbf{Action taken by agent:} \texttt{\{action\}} \\
            \textbf{Reward:} \texttt{\{reward\}} \\
            \textbf{Next state:} \texttt{\{next\_state\}} \\
            \textbf{Collision Flag:} \texttt{\{collision\_flag\}} \\
            \textbf{Lane Index:} \texttt{\{lane\_index\}} \\
            \textbf{Principal Components:} \textbf{PC1} = \texttt{\{pc1\}}, \textbf{PC2} = \texttt{\{pc2\}}, \textbf{PC3} = \texttt{\{pc3\}} \\

            \vspace{0.1cm}

            \textbf{Your Task:}\\
            Analyze each row and provide feedback based on the following criteria: \\
            
            \textbf{1. Action Effectiveness Evaluation:} \\
            a. If the action is effective, respond with: \texttt{llm\_score\_1 = +2} \\
            b. If the action is ineffective, respond with: \texttt{llm\_score\_1 = -2} \\
            c. If the action has no effect, respond with: \texttt{llm\_score\_1 = 0} \\[3pt]
            
            \textbf{2. Reward Appropriateness Evaluation:} \\
            a. If the reward is appropriate, respond with: \texttt{llm\_score\_2 = +1} \\
            b. If the reward is inappropriate, respond with: \texttt{llm\_score\_2 = -1} \\[3pt]

            \vspace{0.1cm}
            \textbf{Expected Format: JSON object containing the following:}\\
            - \textbf{Justification for the decision} \\
            - \textbf{Action Effectiveness Evaluation:} \texttt{<llm\_score\_1>} \\
            - \textbf{Reward Appropriateness Evaluation:} \texttt{<llm\_score\_2>} \\
            \hline
        \end{tabular}
        }
    \end{minipage}%
    \hspace{1.2cm} 
    \begin{minipage}{0.45\textwidth}
        \scalebox{0.66}{
        \renewcommand{\arraystretch}{1.3}
        \begin{tabular}{|p{7cm}|}
            \hline
            \rowcolor{green!30}
            \textbf{LLM $\rightarrow$ Output} \\
            \hline
            \textbf{Analysis:} The action taken by the agent in the fourth lane (Lane Index = 3) led to a collision (Collision flag = 1). Since no explicit control action is specified in this dataset, we infer its effect based on the context. It's reasonable to assume that the action was ineffective in avoiding the collision. The agent could have changed lanes or adjusted speed to prevent the crash.\\
            
            \vspace{0.1cm}
            
            \textbf{Justification:} The agent did not take any action to avoid the collision, indicating an ineffective choice. \\
            
            \vspace{0.1cm}
            
            \textbf{Action Effectiveness Evaluation:} \texttt{llm\_score\_1 = -2} \\
            
            \vspace{0.1cm}
            
            \textbf{Reward Analysis:} The reward given (0.07520053663809241) is positive, suggesting some progress or safety improvement, despite the ineffective action. However, given that a collision occurred, it seems unusual for the reward to be positive. A negative reward would have been more appropriate to discourage such actions in future steps.\\
            
            \vspace{0.1cm}
            
            \textbf{Reward Appropriateness Evaluation:} \texttt{llm\_score\_2 = -1} \\
            \hline
        \end{tabular}
        }
    \end{minipage}
    \caption{The table demonstrates the prompting and output process for a large language model (LLM). The input consists of specific data, including episode number, time step, agent's state, action, reward, collision flag, lane index, and principal components. The output evaluates the action effectiveness and reward appropriateness based on predefined criteria.}
    \label{Tab:Prompt_LLM-D}
\end{table}
\subsubsection{Large Language Model - Direct (LLM-D)}
In this case, we use a LLM (off-the-shelf ) to provide feedback for reward shaping. We do not finetune or train the LLM in any way and use it directly, integrating with our framework via prompts. The LLM acts as a proxy for direct human feedback, offering feedback on agent actions and rewards based on structured trajectory data. 
The LLM receives a data frame where each row contains the episode number, time step, state, action, reward, next state, collision flag, and lane index. The LLM generates feedback based on the state-action pair at each time step, evaluating action effectiveness and reward appropriateness. It categorizes actions as effective, ineffective, or neutral, with adjustments of \(+0.5\), \(-0.5\), or \(0\), and assesses reward appropriateness (+0.2 or -0.2). 

In this context, the LLM generates feedback by processing structured input with the task description or \textit{prompt} (see Table ~\ref{Tab:Prompt_LLM-D}), defined as:

\[
P(y \mid x, t) = f_\sigma(x, t)
\]

Where \( P(y \mid x, t) \) represents the probability distribution of the predicted output \( y \), given the input \( x \) and task description \( t \), with \( x \) being the input text (structured trajectory data), \( t \) as the task description or prompt (feedback evaluation task), and \( y \) as the predicted output (feedback on action and reward). The function \( f_\sigma(x, t) \) represents the learned function of the LLM with parameters \( \sigma \), which processes the input and task description to generate the output.
The reward at each time step is then augmented by adding the LLM feedback to the intrinsic reward from the environment:

\[
\tilde{r}_t = r_t + \zeta_{LLM-D}(s,a;\sigma)
\]

Where \( \tilde{r}_t \) is the augmented reward at time step \( t \), \( r_t \) is the intrinsic reward from the environment, and $\zeta_{LLM-D}(s,a;\sigma)$ is the feedback provided by the LLM at time step \( t \), including action evaluation and reward appropriateness adjustments.

%
%
\begin{table}[h!]
    \normalsize
    \centering
    \scalebox{0.56}{
    \renewcommand{\arraystretch}{1.3}
    \begin{tabular}{|p{18cm}|}
        \hline
        \rowcolor{red!30}
        \textbf{Prompt $\rightarrow$ LLM} \\
        \hline
        \textbf{Data details:} \\
        \textbf{Episode:} \{episode\_num\}, \textbf{Time Step:} \{time\_step\}, \textbf{State:} \{state\}, \textbf{Action:} \{action\}, \\
        \textbf{Reward:} \{reward\}, \textbf{Next state:} \{next\_state\}, \textbf{Collision Flag:} \{collision\_flag\}, \textbf{Lane Index:} \{lane\_index\}, \\
        \textbf{Adjusted score:} \{adjusted\_score\}, \textbf{PC1 (Speed, Lane Changes, Collision Influence):} \{pc1\}, \\
        \textbf{PC2 (Reward Variation, Smooth Driving):} \{pc2\}, \textbf{PC3 (Trajectory Consistency, Stability):} \{pc3\} \\
        \textbf{Your Task: JSON objects of the following:} \\
        \quad \textbf{- VERDICT} \\
        \quad \textbf{- If the adjusted score is correct, respond with:} \\
        \quad \texttt{"Correct score allotted"} \\
        \quad \texttt{llm\_score = \{adjusted\_score\}} \\
        \quad \textbf{- If the adjusted score is biased, respond with:} \\
        \quad \texttt{"Biased score allotted"} \\
        \quad \textbf{- JUSTIFICATION} \\
        \quad \texttt{llm\_score = Suggested appropriate score.} \\
        \hline
    \end{tabular}}
    
    \vspace{0.2cm} 
    
    \normalsize
    \centering
    \scalebox{0.57}{
    \renewcommand{\arraystretch}{1.3}
    \begin{tabular}{|p{18cm}|}
        \hline
        \rowcolor{yellow!30}
        \textbf{LLM $\rightarrow$ Output} \\
        \hline
        \textbf{VERDICT:} BIASED SCORE ALLOTTED\\
        \textbf{Justification:} The presence of a Collision Flag indicates an undesirable event, which has not been penalized in the Adjusted Score (4.0). Although the reward received (0.066) suggests that the agent's action yielded a short-term gain, the llm score(-1.5) penalizes it and emphasizes safety over immediate outcomes. The principal components reinforce this view: PC3 highlights poor trajectory-level consistency, whereas PC1 reflects acceptable control over speed and lane changes.\\
        \textbf{Adjusted Score:} \\
        \quad \texttt{"Biased score allotted"}\\
        \quad \texttt{llm\_score = -1.5}\\
        \textbf{Considerations:} \\
        1. Adjusted Score reflects the human feedback in the form of a composite value that integrates lane changes, collision avoidance, and speed optimization.\\
        2. Reward (0.066) is positive, showing-term beneficial outcome .\\
        3. PC1 suggests acceptable control over speed and lane changes.\\
        4. PC2 reflects moderate stability of reward variation across steps.\\
        5. PC3 highlights poor trajectory-level consistency, alligning with the observed collision and framented beh\\
        \hline
    \end{tabular}}
    
    \caption{Examples on how LLM in the \textit{LLM-HFBF} approach evaluates the input data and adjusts the reward score. The input includes details such as episode number, time step, state, action, reward, collision flag, and compressed representation in form of principal components (PC1, PC2, PC3). In the output, the LLM identifies that the provided adjusted score (4.0) does not align with the agent's actions as reflected in the principal components. It flags the score as biased and adjusts it to -1.5, suggesting a more appropriate score based on reinforcement learning principles. This adjusted score encourages continued learning while acknowledging areas for improvement.}
    \label{Tab:Prompt_LLM-HFBF}
\end{table}

This feedback mechanism allows for structured reward shaping, effectively replacing direct human input with LLM-generated feedback, leveraging zero-shot learning to adapt to new situations without explicit retraining.

We choose \textbf{Mistral 7.2B} \cite{jiang2023mistral7b} model 
due to its efficient architecture, strong contextual understanding, and relatively low computational cost. 
In the context of environments like Highway, its ability to process structured trajectory descriptions enables effective evaluation of lane changes, collision avoidance, and speed optimization, while its smaller size ensures faster inference for real-time feedback.
However, limitations include potential biases, reduced reasoning depth, and occasional inconsistencies. Larger or fine-tuned models could improve robustness and accuracy.

\subsubsection{LLM Aided Human Feedback Bias Flagging (LLM-HFBF)} The LLM-HFBF approach represents our hybrid framework that leverages LLMs to detect and correct potential biases in human feedback (see Fig.~\ref{fig:feedback_strategies} for the LLM-HFBF workflow and  Table~\ref{Tab:Prompt_LLM-HFBF} shows the prompt and corresponding LLM feedback). This approach can be formalized as follows:
Let
\[
\mathcal{D} = \{ (s_t, a_t, r_t, \hat{r}_t^{HF}) \}_{t=0}^{T}
\]
be the trajectory set generated from an RL policy $\pi_{\phi}$, where $s_t$, $a_t$, $r_t$, and $\hat{r}_t^{HF}$ represent the state, action, intrinsic reward, and human feedback-based shaped reward at time $t$, respectively.
The human feedback-based shaped reward $\hat{r}_t^{HF}$ is obtained using the biased HF-D approach:
\[
\hat{r}_t^{HF} = \zeta_{\text{HF-D}}(s_t, a_t, p)
\]
where $p \in \{\text{AGG}\}$.

To detect and correct potential biases in $\hat{r}_t^{HF}$, we leverage the LLM through a bias detection function:
\[
\delta_{\text{LLM}}(s_t, a_t, \hat{r}_t^{HF}, \boldsymbol{\omega}_t; \sigma) =
\begin{cases}
\hat{r}_t^{HF}, & \text{if bias not detected} \\
\hat{r}_t^{LLM}, & \text{if bias detected}
\end{cases}
\]
where $\boldsymbol{\omega}_t = [PC1_t, PC2_t, PC3_t]$ represents the principal components of the trajectory at time $t$, which capture Speed/Lane Changes/Collision Influence (PC1), Reward Variation/Smooth Driving (PC2), and Trajectory Consistency/Stability (PC3). The parameter $\sigma$ represents the LLM parameters, and $\hat{r}_t^{LLM}$ is the corrected reward suggested by the LLM.

The objective representation of trajectory-level agent behavior in the LLM-HFBF approach is derived from the following raw features: (1) State, (2) Action, (3) Reward, (4) Collision flag, (5) Lane index, and (6) Adjusted rewards (computed via biased reward shaping). To extract the most informative structure from this high-dimensional feature space, Principal Component Analysis (PCA) was applied. The first three principal components (PCs) capture 95.17\% of the total variance, with PC1 accounting for 45.39\%, PC2 for 27.92\%, and PC3 for 21.86\%. This dimensionality reduction facilitates efficient clustering and analysis by preserving essential behavioral patterns—including speed regulation, lane-change strategies, collision risk, and overall trajectory consistency—while filtering out redundant correlations across features.

The LLM generates the bias detection and correction by processing a structured prompt that includes the state-action pair, the human-suggested reward, and the principal components:
\[
f_{\sigma}(s_t, a_t, r_t, s_{t+1}, c_t, l_t, \hat{r}_t^{HF}, \boldsymbol{\omega}_t) \rightarrow (b_t, \hat{r}_t^{LLM})
\]
where $c_t$ is the collision flag, $l_t$ is the lane index, $b_t \in \{\text{True}, \text{False}\}$ indicates whether bias was detected, and $\hat{r}_t^{LLM}$ is the corrected reward if bias is detected.

The final augmented reward for the LLM-HFBF approach is:
\[
\tilde{r}_t = r_t + \zeta_{\text{LLM-HFBF}}(s_t, a_t, \hat{r}_t^{HF}, \boldsymbol{\omega}_t; \sigma)
\]
where
\[
\zeta_{\text{LLM-HFBF}}(s_t, a_t, \hat{r}_t^{HF}, \boldsymbol{\omega}_t; \sigma) = \delta_{\text{LLM}}(s_t, a_t, \hat{r}_t^{HF}, \boldsymbol{\omega}_t; \sigma)
\]
This approach enables selective intervention, where the LLM only modifies human feedback when bias is detected based on objective trajectory characteristics captured by the principal components. For instance, if the adjusted score (human feedback) is 0.0 but the principal components indicate beneficial agent behavior (e.g., no collision, appropriate lane positioning, etc.), the LLM can flag this as biased and suggest an improved score (e.g., 0.3). 
The corrected trajectories
\[
\mathcal{D}^{'} = \{ (s_t, a_t, \tilde{r}_t) \}_{t=0}^{T}
\]
are then used for policy optimization.

\section{Experiments and Analysis}

\subsection{Experiment Setup}\label{Sec:Env_def}
Below we describe the different RL environments, custom edge case scenarios, and metrics we have used to evaluate and compare the performance of the different FIS.
\subsubsection*{A.1. Highway (Highway-v0)}
\textit{A.1. Default environment: } 
It is a multi-lane driving scenario with four lanes and a total of 50 vehicles, where the agent controls one vehicle. The agent’s initial lane ID is unspecified, allowing flexible starting positions. The reward function includes a collision penalty of -1, no direct reward for lane changes (0 reward), and an incentive for staying in the right lane with a 0.1 reward, given by
\[
r =
\begin{cases}  
-1, & \text{collision penalty} \\  
0, & \text{lane change} \\  
0.1, & \text{staying in the right lane} \\  
0.4, & \text{speed } \in [20, 30] \text{ m/s}  
\end{cases}
\]
High-speed driving is encouraged with a reward of 0.4 for maintaining speeds within the 20 to 30 m/s range. Other vehicles follow the Intelligent Driver Model (IDM) for realistic traffic behavior.
\newline
\subsubsection*{A.2. Custom Edge Case Scenarios: }
    \begin{enumerate}
        \item \textit{Case 1: Congested Lane}\newline
        The rightmost lane, typically rewarded in highway-v0, is heavily congested with slower-moving vehicles (speeds between 15-20 m/s). The adjacent lanes are moderately occupied with faster-moving traffic (25-30 m/s). The ego vehicle starts in the rightmost lane at 25 m/s and must navigate around congestion to maintain high speed while avoiding collisions. 
        The highway environment’s duration parameter, which sets the max timesteps per episode if not terminated, is set to 40 by default.

        \item \textit{Case 2: Slow-Moving Obstacle in Middle Lane} \newline
        This scenario introduces a slow-moving vehicle (speed: 15 m/s) positioned in the middle lane, directly in the ego vehicle’s path. Adjacent lanes have moderately dense traffic traveling at 25-30 m/s. The ego vehicle begins in the middle lane at 25 m/s and must make precise lane changes to overtake the obstacle safely and return to the middle lane for optimal reward. 
        
    \end{enumerate}

\subsubsection*{A.2. Reacher (Reacher-v5)}
Reacher-v5 in Gymnasium is a robotic control environment where a two-jointed arm must reach a target in a 2D plane. The state space is 11-dimensional, including joint angles, angular velocities, and relative positions of the target and end-effector. The action space is continuous (2D), representing torques applied to the joints. The reward function is given as
\[
r = -w_{\text{near}} \| p_{\text{eff}} - p_{\text{tgt}} \|_2 - w_{\text{ctrl}} \| a \|_2^2
\]

where \textbf{$p_{eff}$} (end-effector position) is the 2D coordinate of the fingertip of the robotic arm, \textbf{$p_{tgt}$} (target position) is the 2D coordinate of the goal location, and $\| p_{eff} - p_{tgt} \|_2$ is the Euclidean distance between them, which the reward function minimizes to encourage precise reaching, \( w_{\text{near}} = 1.0 \), \( w_{\text{ctrl}} = 0.01 \). Encourages target proximity, penalizes large actions.

\subsection{Metrics Used}
\noindent
To measure the performance of LLM-based reward shaping and human feedback across these behavioral variations, we used the following metrics.
\begin{enumerate}
   
    \item \textit{Average Episodic Reward (AER):} The cumulative reward for a given episode \(e\) is the sum of the rewards over all timesteps within that episode. Let \(R_{e,i}\) represent the reward at timestep \(t\) in episode \(e\), and let \(T_e\) be the total number of timesteps in episode \(e\). Then, the cumulative reward \(R_{C_{e}}\) for episode \(e\) is expressed as:

    \[
    R_{C_{e}} = \sum_{t=1}^{T_e} R_{e,t} 
    \]
    
    The average cumulative reward across all episodes, is then computed as:
    
    \[
    AER = \frac{1}{N} \sum_{e=1}^{E} R_{C_{e}}= \frac{1}{E} \sum_{e=1}^{E} \sum_{t=1}^{T_e} R_{e,t}
    \]
    
    where $R_{C_{e}}$ is the cumulative reward for episode \(e\),
    \(E\) is the total number of episodes,\(T_e\) is the number of timesteps in episode \(e\) and  \(R_{e,t}\) is the reward at timestep \(t\) in episode \(e\).
    \vspace{0.1cm}

    \item \textit{Average Terminate Time (ATT):} is the average number of timesteps before termination (either due to a collision or reaching the default maximum timestep) across 
    $E$ episodes in the Highway environment. This metric measures how long the vehicle drives before either colliding or reaching the end of the episode. For each episode \( e \), the time to collision is the number of timesteps until either a collision or the maximum timestep (default is 40) is reached.
    
    \[
    \text{ATT} = \frac{1}{E} \sum_{e=1}^{E} T_{e}^{L}
    \]
    
    where \( T_{e}^{L} \) is the number of timesteps in episode \( e \) before a episode terminates.
    
    \item \textit{Feedback Misalignment (FMA):} This metric assesses deviations from optimal driving behavior based on trajectory data. It considers four key factors: unnecessary lane changes, suboptimal actions, collision risks, and speed variations.

    \[
    \text{FMA} = \sum_{t=1}^{T} \left[ \lambda_1 \mathbf{1}(\Delta \text{Lane}_t) + \lambda_2 \mathbf{1}(\text{Reward}_t < \theta) \right. \quad \\
    \left. + \lambda_3 \mathbf{1}(\text{Collision}_t) + \lambda_4 |\Delta \text{Speed}_t| \right]
    \]

    where \( \Delta \text{Lane}_t = (\text{Lane Index}_t \neq \text{Lane Index}_{t-1}) \) counts lane changes,  
    \( \text{Reward}_t < \theta \) flags suboptimal actions,  
    \( \text{Collision}_t \) is 1 if a collision occurs,  
    \( \Delta \text{Speed}_t = \text{Speed}_t - \text{Speed}_{t-1} \) measures acceleration fluctuations,  
    and \( \lambda_1, \lambda_2, \lambda_3, \lambda_4 \) are weighting factors based on importance.

\end{enumerate}
Among the three metrics, FMA and ATT are environment-specific (to Highway), while  AER is environment invariant. 
In each of these environments, we assessed the learning speed, stability, and overall performance of agents using LLM-based reward shaping compared to traditional human feedback methods.

\subsection{Results and Discussion}

Our experimental results (see Table~\ref{tab:default_highway_environment}) clearly demonstrate that even minor biases in human feedback can substantially degrade the performance of the reinforcement learning agent. In the default highway environment, the unbiased approaches (HF-RSM, HF-D, LLM-D, LLM-HFBF) consistently achieved near-optimal performance. The Average Episodic Reward (AER) for these methods ranged from 27.608 to 28.472. The Feedback Misalignment (FMA) was 0.6 for HF-RSM and 0.0 for HF-D, LLM-D, and LLM-HFBF. All of these unbiased methods achieved a maximum Average Terminate Time (ATT) of 40.0.  

In contrast, biased human feedback led to significant performance deterioration. The biased HF-D approaches showed the most severe performance drops. With aggressive bias (AGG), the AER dropped to 5.863, FMA increased to 5.6, and ATT fell to 7.0. Reckless adaptive bias (RAD) caused the AER to drop further to 1.840, FMA to reach 5.0, and ATT to decline to 3.0. This suggests that direct biased human feedback introduces substantial misalignment with optimal driving behavior, resulting in more frequent collisions and shorter episode durations.
.\\

\begin{table}[t]
    \centering
    \scalebox{0.9}{
    \begin{tabular}{|l|c|c|c|}
    \hline
    \textbf{FIS} & \textbf{FMA} & \textbf{ATT} & \textbf{AER (Highway) } \\ \hline
    HF-RSM         & 0.6& 40.0& 27.608\\ \hline
    HF-D            & 0.0  & 40.0& 28.472\\ \hline
    \textit{biased} HF-RSM (AGG) & 0.6& 40.0& 29.431\\ \hline
    \textit{biased} HF-D (AGG)      & 5.6& 7.0& 5.863\\ \hline
    \textit{biased} HF-RSM(RAD)& 1.6& 34.4& 25.448\\ \hline
    \textit{biased} HF-D(RAD)& 5.0& 3.0& 1.840\\ \hline
    LLM-D           & 0.0  & 40.0& 28.472 \\ \hline
    LLM-HFBF & 0.0  & 40.0& 28.472 \\ \hline
    \end{tabular}

}
    \caption{Performance of different FIS, on default Highway Environment. The policy is trained using PPO\cite{schulman2017proximal} from StableBaselines3 (SB3) implementation\cite{stable-baselines3}, for 0.01 Million timesteps. }
    \label{tab:default_highway_environment}
\end{table}

The Human Feedback with Reward Shaping Model (HF-RSM) demonstrated greater resilience to bias compared to direct human feedback (HF-D). Even when biased, HF-RSM maintained relatively better performance with an AER of \textcolor{black}{25.448}, lower FMA of \textcolor{black}{1.6}, and higher ATT of \textcolor{black}{34.4}. This indicates that surrogate models can partially mitigate the effects of bias through their generalization capabilities, acting as a buffer between potentially skewed human evaluations and the agent's learning process. The LLM-based direct feedback approach (LLM-D) exhibited remarkable effectiveness, achieving optimal performance metrics (AER: 28.472, FMA: 0, ATT: 40.0) equivalent to unbiased human feedback. This validates the hypothesis that large language models can effectively replace human feedback without requiring task-specific training. Despite having no prior specialized training on highway navigation tasks, the LLM provided feedback comparable to unbiased human input.\newline

\begin{table}[htbp]
    \centering
    \scalebox{0.9}{
    \begin{tabular}{|c|l|c|c|c|}
        \hline
         & \textbf{FIS} & \textbf{FMA } & \textbf{ATT} & \textbf{AER (Highway)} \\
        \hline
        \multirow{4}{*}{\rotatebox{90}{\textbf{Case 1}}} & HF-D & 5.0 & 20.2& 14.174 \\
        \cline{2-5}
        & \textit{biased} HF-D (AGG) & 5.8& 5.4& 4.332\\
        \cline{2-5}
        & \textit{biased} HF-D (RAD)& 8.8& 7.4& 5.856\\
        \cline{2-5}
        & LLM-D & 5.0 & 20.2& 14.174 \\
        \hline
        \multirow{4}{*}{\rotatebox{90}{\textbf{Case 2 }}} & HF-D & 5.0 & 19.6& 13.299 \\
        \cline{2-5}
        & \textit{biased} HF-D (AGG) & 6.6 & 7.8& 6.812\\
        \cline{2-5}
        & \textit{biased} HF-D (RAD)& 8.0& 6.0& 4.682\\
        \cline{2-5}
        & LLM-D & 5.0 & 19.6& 13.299 \\
        \hline
    \end{tabular}
    }
    \caption{Comparison of performance of different FIS on two edge case scenarios of the Highway environment.}
    \label{tab:edge_case_comparison}
\end{table}

The edge case experiments in Table~\ref{tab:edge_case_comparison} further highlight the robustness of LLM-based approaches. In both challenging scenarios, LLM-D matched the performance of unbiased HF-D (AER: 14.174 in Case 1, 13.299 in Case 2), while biased feedback approaches continued to produce suboptimal results. In Case 1, biased HF-D with aggressive bias (AGG) had an AER of \textcolor{black}{4.332} with FMA of \textcolor{black}{5.8}, while \textcolor{black}{reckless adaptive bias (RAD)} had an AER of \textcolor{black}{5.856} with FMA of \textcolor{black}{8.8}. In the more challenging Case 2, biased HF-D \textcolor{black}{again performed poorly}, with aggressive bias resulting in an AER of \textcolor{black}{6.812} and FMA of \textcolor{black}{6.6}, while \textcolor{black}{reckless adaptive bias} led to the lowest performance with an AER of just \textcolor{black}{4.682} and FMA of \textcolor{black}{8.0}. This demonstrates how rule-based approaches struggle with contextual adaptation, particularly when optimal behavior requires deviating from rigid predefined rules.\newline

\begin{table}[h]
    \centering
    \begin{tabular}{|l|c|}
        \hline
        \textbf{FIS} & \textbf{AER (Reacher)} \\
        \hline
        HF-D & -38.210\\
        \hline
        \textit{biased} HF-D (AGG) & -80.336\\
        \hline
        \textit{biased }HF-D (RAD)& -30.009\\
        \hline
        LLM-D & -11.637\\
        \hline
        LLM-HFBF & -12.164\\
        \hline
    \end{tabular}
    \caption{Comparison of performances of different FIS on the Reacher environment (The policy is trained using PPO \cite{schulman2017proximal} from SB3 \cite{stable-baselines3} implementation, for 0.2 Million time-steps). The HF-D biased and unbiased rules and the LLM prompt structure are conceptually similar to that of the Highway case as elaborated in Sec.~\ref{Sec: HF-D}. }
    \label{tab:Reacher_perf}
\end{table}
To further assess the robustness and generalizability of LLM-based approaches, we tested them in a second continuous control environment, \textbf{Reacher}. The results, shown in Table~\ref{tab:Reacher_perf}, reinforce the trend observed in the Highway environment. LLM-based approaches continued to match the performance of unbiased HF-D, while biased feedback approaches resulted in substantial performance degradation.

\begin{figure}[h!]
    \centering
    \includegraphics[width=0.9\linewidth]{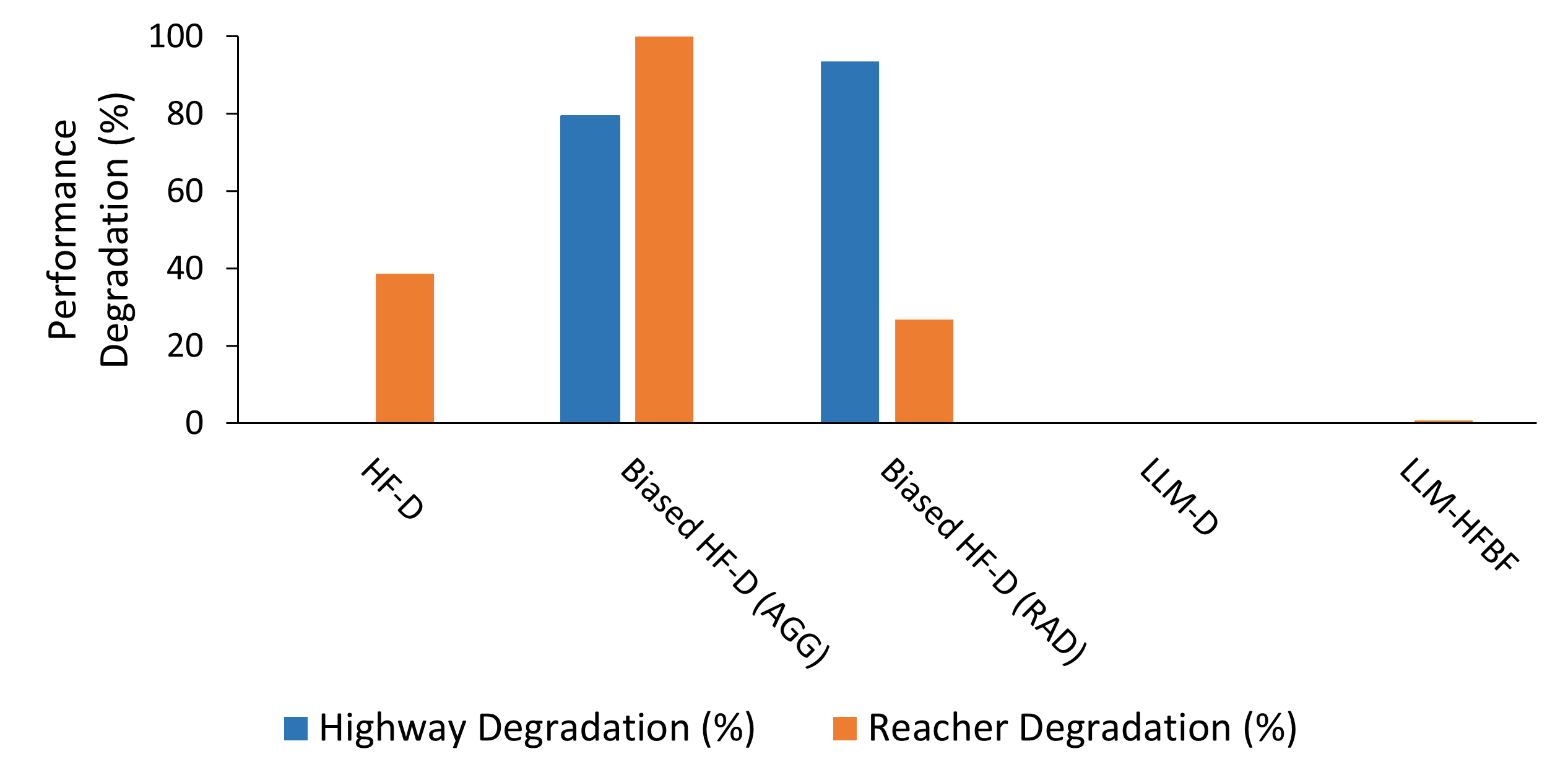}
    \caption{Performance Degradation Due to Bias Across Environments: 
    compares the performance degradation of different feedback strategies 
    (LLM-D, LLM-HFBF, and two biased HF-D variants) relative to the unbiased human 
    feedback (HF-D) baseline. The values on the y-axis represent the percentage 
    decrease from the optimal performance achieved by HF-D in each environment, 
    with lower percentages indicating less degradation. The data from the Highway 
    and Reacher environments are shown side-by-side. The optimal baseline Average 
    Episodic Reward (AER) for these calculations is 28.472 for Highway and -11.637 
    for Reacher. The final values are normalized to a 0--100 scale using Min--Max 
    normalization
   }
    \label{fig:cross_env}
\end{figure}

Additionally to compare the performance of the LLM based approaches across environments, we present normalized comparative plots in Fig.~\ref{fig:cross_env}. LLM-based approaches maintain stable performance across environments, highlighting their consistency. Aggressive bias (AGG) leads to a significant \textcolor{black}{$\sim$79\%} drop in Highway and \textcolor{black}{$\sim$110\% }in Reacher, while \textcolor{black}{reckless adaptive bias (RAD)} causes the most severe degradation—\textcolor{black}{$\sim$94\%} in Highway and \textcolor{black}{$\sim$21\%} in Reacher. LLM-HFBF performs \textcolor{black}{nearly identical to LLM-D in both environments, showing no gap for the Highway environment and a marginal difference for the Reacher environment.} The consistent impact of bias across environments reinforces the robustness of these findings, confirming the reliability of performance trends observed in different settings.

These findings have key implications for Human-In-The-Loop Reinforcement Learning. First, detecting and mitigating human biases is crucial to maintaining optimal performance. Second, large language models can provide effective feedback without domain-specific training, enabling more scalable RL systems. Third, integrating human expertise with LLM-driven bias detection offers a balanced approach, allowing selective human intervention only when needed. Finally, the success of LLM-based feedback in continuous control tasks suggests its broader applicability across diverse RL domains.\newline

\section{Conclusion}
In this work we successfully demonstrate that zero-shot LLMs can effectively replace human feedback for reward shaping in continuous control tasks. Through experiments, we showed that biased human feedback can significantly degrade RL agent performance, with reductions of up to 94\% in average episodic reward. In contrast, the proposed LLM-based methods (LLM-D and LLM-HFBF) maintain performance levels comparable to unbiased feedback, even in challenging edge-case scenarios. Moreover the LLM-HFBF hybrid framework offers a promising solution that addresses both the inherent biases in human supervision and the limitations of LLMs, creating a more balanced and reliable system for HITL-RL applications.

\section{Limitations and Future work}
Despite promising results, several important limitations warrant further investigation. First, our current implementation requires trajectory data frames to be relatively small due to LLM context window constraints, limiting the historical information available for feedback generation. Second, obtaining feedback from LLMs incurs significant computational costs, requiring GPU resources that may not be available in all application contexts, especially for real-time decision-making. Third, our approach depends heavily on iterative and precise prompt design, which requires expertise and may need adjustment across different environments. Additional limitations include environmental complexity beyond highway navigation, the need to explore different LLM architectures, and methods for real-time LLM-based feedback. Future work will address these limitations and explore applications to a broader range of continuous control tasks, including robotic manipulation and autonomous driving in diverse scenarios.

\bibliographystyle{elsarticle-num}  
\bibliography{reference}          

\end{document}